# Shape Classification using Approximately Convex Segment Features

**BIMAL KUMAR RAY**
School of Computer Science Engineering and Information Systems
Vellore Institute of Technology, Vellore – 632014, INDIA

**ABSTRACT** The existing object classification techniques based on descriptive features rely on object alignment to compute the similarity of objects for classification. This paper replaces the necessity of object alignment through sorting of feature. The object boundary is normalized and segmented into approximately convex segments and the segments are then sorted in descending order of their length. The segment length, number of extreme points in segments, area of segments, the base and the width of the segments – a bag of features – is used to measure the similarity between image boundaries. The proposed method is tested on datasets and acceptable results are observed.

## I INTRODUCTION

A number of studies based on either descriptive features or learned features have been carried out toward shape classification. The concept of shape context [1] and a variation [2] of it, using height function in classification [3], inner distance for classification [4], angular pattern and binary angular pattern [5], hierarchical projective invariant contexts [6] are based on descriptive features whereas as deep learning based classification techniques such as vision transformer [7], CNN [8], [9], Sketch-a-Net [10], PointNet (for 3D) [11] rely on learning shape features through extensive training of the network and tuning the hyper parameters in the process. The descriptive features based approaches do not take into account the texture and color information of the image whereas deep learning based approaches rely heavily on texture and color information though they are also found to be successful to classify silhouettes. Qi Jia et al. [12] devised a cartoon character recognition scheme using multi-level shape context and positional encoding maintaining rotation robustness. Xiaohan Yu et al. [13] proposed a deterministic analytical approach using Lie algebra for shape classification with distance maps, local binary patterns and image intensity as descriptors of region.

This paper proposes a feature-based silhouette object classification technique that replaces the task of object alignment by sorting object boundary segments in descending order of their length; because longer segments are perceptually more dominant than the smaller segments. The object boundary is segmented, sorted in descending order based on their length and segment features namely segment length, number of extreme points (maxima and minima curvature points), area of segment and its base and width are used to compute the similarity between two objects with respect to their boundary characteristics.

The proposed technique is based on a bag of features of curve segments. It can operate successfully on object with curved boundaries and silhouette but cannot be successful on objects with texture. It is not possible for the proposed scheme to recognize leaves from a leave database because leaves have vein structures which, apart from its boundaries, are the defining features of a leaf.

## II METHODOLOGY

A closed digital boundary can be described by a sequence of coordinates as in
$C = \{(x_i, y_i), i = 1, \dots n \text{ and } i = (i \pm n) mod(n)\}$.
The center of mass of the curve is defined by $x_c = \frac{\sum_1^n x_i}{n}$ and $y_c = \frac{\sum_1^n y_i}{n}$. The input data are normalized using the center of mass of the object boundary as its translation parameter and the reciprocal of the object perimeter as the object's scale parameter. The object boundary is segmented using a merging technique followed by detection of approximately convex segments of the object boundary. The merging process involves looking for the longest line segment with deviation from the curve constrained within a threshold. Starting from an arbitrary point on the curve boundary; points are merged one after another as long as the line segment joining the fixed point to the currently visiting point deviates from the boundary maximally by a threshold determined by the object scale. This process basically looks for the longest possible line segment between a fixed point and the currently visiting point within an error tolerance which is not a user input; rather it is determined by the reciprocal of the perimeter of the curve boundary (called object scale). The maximal deviation of a segment is defined by the maximal distance to the line segment joining the fixed point and the current point. If a point on the curve falls outside the segment then the distance is measured by its distance from the nearest end point of the scanning line segment otherwise, it is the orthogonal distance of the point from the current segment.

If $p_i(x_i, y_i)$ be a starting point and $p_j(x_j, y_j)$ be the currently visiting point then the maximal deviation of the curve segment from the line segment $L_{ij}$ joining the points $p_i$ and $p_j$ is the maximum perpendicular distance of a point $p_k$ ($i < k < j$) on the curve, if the foot of the perpendicular from the point at the maximal distance falls on the line segment $L_{ij}$. If the point $p_k$ ($i < k < j$) has the foot of its perpendicular onto the left (or right) of the line segment $L_{ij}$ then the distance of $p_k$ is determined by its Euclidean distance from the point $p_i(x_i, y_i)$ (or $p_j(x_i, y_i)$). As it is evident here, the distance to a point from the line segment $L_{ij}$ is used as a metric to compute the maximal deviation of the curve from the line segment $L_{ij}$.

Whenever the maximal deviation exceeds the threshold; a new starting point is detected. This process is continued through the entire curve until a new starting point coincides with one of the landmark points already generated by the scan. This completes one pass through the curve. The subsequent passes are carried out by incrementing the threshold by the value of the object scale. It may be noted here that user intervention for threshold value is replaced by automatic threshold which is initialized by object scale and is incremented by the same. The iterative process is continued as long as the sum of square of errors of an approximation remains restricted within a measure defined by the product of the square of threshold and the reciprocal of the current compression ratio (defined by the ratio of the number of points on the curve to the number of landmark points). This is a heuristic for terminating the iteration wherein the upper bound imposed on the sum of square of errors leads to avoidance of too much distortion in the approximation with respect to the curve. As the threshold is increased; the number of vertices decreases and consequently, the approximation error increases. The heuristic is used to maintain the approximation error within a tolerance and when the tolerance is broken; the iteration is terminated and the maximum value of the threshold is generated. This threshold is used in the subsequent deletion phases. The other possible heuristics were also tried with but this one is found to produce better results than the others.

After the termination of the iterative process; the last value of the threshold (called error tolerance) used in the sequential scanning process is now used to merge the boundary further through a three-phase deletion process so as to eliminate pseudo landmark points from the curve. The first phase of deletion involves determination of the segment of the curve with the minimum deviation from the curve. This segment is called the weakest segment and it is merged with its adjacent segments if the maximum approximation error that results after merging the segments does not exceed the error tolerance. The maximum of the distance of a point on a segment of the curve is the highest deviation of the segment to the line segment joining the end points of the segment. The process is carried out for every weakest segment in iteration as long as permissible by the error tolerance.

The second phase of deletion is similar to the first phase but here the approximation is relaxed using the sum of error tolerance and a multiple of the object scale as the threshold for deletion. This threshold is, as it may be observed, computed based on error tolerance, object scale and a hyper parameter which is supposed to be tuned for a data set. The third phase of deletion has another hyper parameter as a threshold on the cosine of the angle at the adjacent segments of the approximation. The cosine of the angle between two segments is defined by ratio of the scalar product of the vectors originating from the common point of the segments and the product of their length. If two adjacent segments are determined by the three consecutive segment points $q_l$, $q_m$, and $q_n$ ($either\ l < m < n\ or\ l > m > n$) then the cosine of the inclusive angle of the segments is defined by $\cos \theta = \frac{\overrightarrow{q_m q_l} \cdot \overrightarrow{q_m q_n}}{|\overrightarrow{q_m q_l}| \cdot |\overrightarrow{q_m q_n}|}$. The segment with the minimum cosine value is identified and is merged with its adjacent segments if the cosine value does not exceed the value of a hyper parameter. The process is repeated at every bent as long as the cosine at the bent does not exceed the hyper parameter. The merging using sequential scan followed by deletion of pseudo landmark points generates an approximation some of which from MPEG7 dataset are shown in the Figure 1.

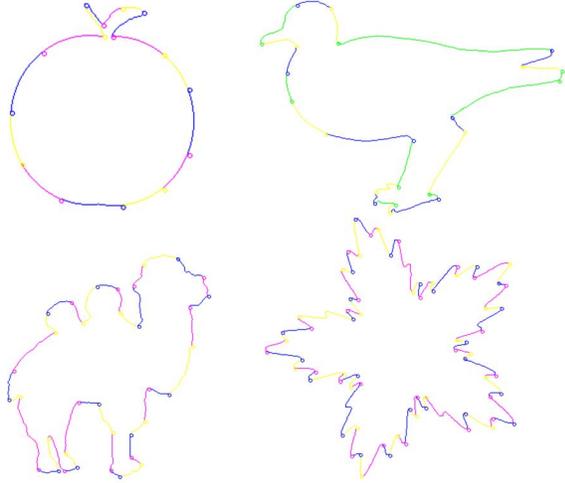

**FIGURE 1** Boundary segmentation through scanning and multi stage deletion. Segments are coded with different color

This approximation is now searched for identifying approximately convex segments on the curve. The vector product is used to identify the concave landmark points. If $S_a, S_b$ and $S_c$ ($a > b > c$ or $c > b > a$) be three consecutive landmark points then the scalar component of the vector product $\overrightarrow{S_b S_a} \times \overrightarrow{S_c S_b}$ determines the concavity of the landmark point $S_b$. If it is negative then the point $S_b$ is a segmentation point on the approximately convex segment representation of the boundary. Some results of approximately convex segments decomposition are shown in Figure 2.

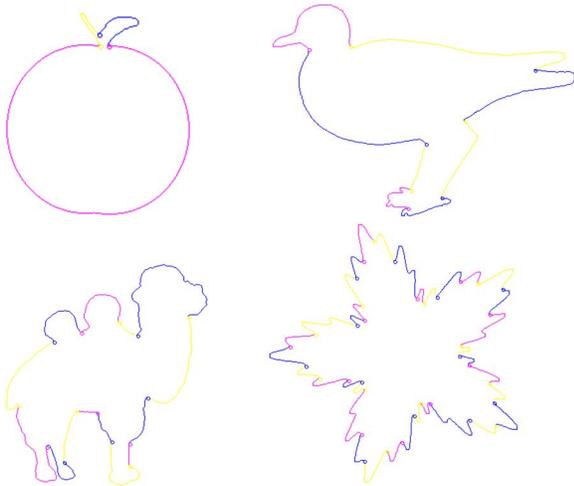

**FIGURE 2** Approximately convex segments of object boundary. Different convex segments are coded with different color

The approximately convex segments decomposition scheme defines a compact representation of the curve boundary as a sequence of approximately convex segments and is used to compute descriptive features of the segments of the curve. The segment size, number of extreme (maximum and minimum of curvature) points on a segment, area of a segment, and the base and width of the segment are computed as their descriptive features.

The literature on object classification and recognition has been found to apply rotation and translation to align one curve boundary with another to facilitate classification and recognition. In this paper, the curve segments are sorted in descending order of their lengths and the mechanism of curve alignment is not used. It is assumed that the sorting process of the segments aligns two different curves if they belong to the same class. The similarity measure between two objects is computed by the sum of the squared difference in the size of the corresponding segments (after sorting), number of extreme curvature points on the segments, base width and height of the segments and the area of the corresponding segments. The segment size is computed as the number points present in the segment. The segment size of the corresponding segments of two objects is a distinguishing feature to isolate one object from another. The number of extreme points too is a distinguishing feature of segments. The extreme points are the points with maximum and minimum curvature. The direct curvature measurement would suffer from the problem of scale. So curvature is not measured directly; rather an alternative way of finding the extreme curvature points is used. The distance of the segment points from the centroid of the curve segment is tracked for its minimum and maximum value and the total number of such local minima and maxima over the segment is taken as the number of extreme points of the segment. The base of a segment is an indicative of how much wide a segment is and its height indicates the protrusion of the segment. The base width is measured by the linear distance between the end points of the segment and the height is the maximum distance of the segment boundary from its base.

It is very likely that two objects, even if they belong to the same class. may have different number of segments. The curve $C_i$ and $C_j$ can have segments $s_i$ and $s_j$ (say) number of segments with $s_i \neq s_j$ and it is assumed that $s = \max(s_i, s_j)$. The minimum is not considered because it results in loss of information. Assume that the length of a segment, its number of extreme points, its area, and its base and height for the $k^{th}$ segment of the $i^{th}$ curve are denoted respectively by $n_{ik}, x_{ik}, a_{ik}, b_{ik}$ and $h_{ik}$. Since the maximum of the number of segments instead of the minimum in a pair of objects is being considered so the value of the descriptive features for the segments of the object with smaller number of segments is taken as zero because of their non-existence in the object. The similarity measure is computed as the sum of squared difference of the descriptive features of two curves after sorting the features in descending order of segment length. The similarity measure between the curves $C_i$ and $C_j$ is computed by
$S_{ij} = \sum_{k=1}^{s}\{(n_{ik} - n_{jk})^2 + (x_{ik} - x_{jk})^2 + (a_{ik} - a_{jk})^2 + (b_{ik} - b_{jk})^2 + (h_{ik} - h_{jk})^2\}$.

## III PERFORMANCE

The procedure of segmentation and classification is implemented in Java using NetBeans IDE 8.2 on a computer with a RAM of 2GB and a CPU speed of 2.40GHz. The performance of the classification scheme is measured using leave-one-out-cross-validation strategy. The MPEG7 dataset and Kimia99 dataset are used for testing the scheme. The performance (accuracy) of the scheme on MPEG7 dataset is found to be 81.86 and on Kimia99 dataset it is 87.88 in terms of accuracy. The execution time on MPEG7 dataset is 55 seconds and that on Kimia99 dataset is one second.

## IV CONCLUSION

The key aspect of this paper, in contrast to the conventional feature-based approaches to shape classification based on object features is to replace object alignment aspect of classification scheme by sorting features of segments of the object. The object boundary is decomposed into approximately convex segments and the segments are sorted based on segment size. A bag of features is used to classify objects. The scheme is tested on two popular datasets and acceptable results are observed.